\documentclass{article}

\usepackage{ifthen}
\newboolean{ack}

\usepackage{microtype}
\usepackage{graphicx}
\usepackage{subfigure}
\usepackage{booktabs} 
\usepackage{breqn}
\usepackage{multicol}
\usepackage{wrapfig}
\usepackage{makecell}
\usepackage[skip=2pt]{caption}

 \usepackage[final,nonatbib]{neurips_bdl2019}  \setboolean{ack}{true}

\usepackage[utf8]{inputenc} 
\usepackage[T1]{fontenc}    
\usepackage{hyperref}       
\usepackage{url}            
\usepackage{booktabs}       
\usepackage{amsfonts}       
\usepackage{nicefrac}       
\usepackage{microtype}      
\usepackage{algorithm}
\usepackage{algorithmic}
\usepackage{enumitem}
\usepackage{textcomp}
\usepackage{color}

\title{Towards calibrated and scalable uncertainty\\representations for neural networks}

\author{Nabeel Seedat$^{1,2,3}$ \qquad \qquad \qquad \textbf{Christopher Kanan}$^{2,4,5}$ \\
$^1$Cornell University \qquad $^2$Cornell Tech \qquad $^3$Shutterstock \\\qquad $^4$Rochester Institute of Technology\qquad $^5$Paige\\
{\texttt \small ns879@cornell.edu \qquad kanan@rit.edu}
}

\begin{document}

\maketitle

\begin{abstract}
For many applications it is critical to know the uncertainty of a neural network's predictions. While a variety of neural network parameter estimation methods have been proposed for uncertainty estimation, they have not been rigorously compared across uncertainty measures. We assess four of these parameter estimation methods to calibrate uncertainty estimation using four different uncertainty measures: entropy, mutual information, aleatoric uncertainty and epistemic uncertainty. We evaluate the calibration of these parameter estimation methods using expected calibration error. Additionally, we propose a novel method of neural network parameter estimation called RECAST, which combines cosine annealing with warm restarts with Stochastic Gradient Langevin Dynamics, capturing more diverse parameter distributions. When benchmarked against mutilated image data, we show that RECAST is well-calibrated and when combined with predictive entropy and epistemic uncertainty it offers the best calibrated measure of uncertainty when compared to recent methods. 
\end{abstract}

\section{Introduction}
Breakthroughs in deep learning have greatly improved the capabilities of neural networks in image understanding~\cite{Krizhevsky2012}, speech recognition~\cite{speech}, self-driving cars~\cite{car}, medical image diagnostics~\cite{medical}, and much more. For sensitive applications, such as self-driving cars and medical image analysis, it is critical for a neural network to provide uncertainty estimates for its decisions. More specifically, the network indicating when it is likely to be incorrect \cite{guo2017calibration}.

Unfortunately, off-the-shelf methods lack this capability~\cite{gal2015dropout,maddoxfast,Pearce2018UncertaintyIN,sri}. While softmax outputs are predictive probabilities, they are not a valid measure for the confidence in a network's predictions. Moreover, these probabilities are often poorly calibrated and result in overconfident predictions~\cite{guo2017calibration,nguyen2015deep,Kendall2017}. 

Bayesian methods offer a principled approach of uncertainty representation in neural networks by representing all network parameters in a probabilistic framework. The challenge associated with Bayesian methods is the high computational cost associated with inference. This has resulted in an inability to scale the solutions. Moreover, while many works focus on inference in Bayesian methods, they often ignore the effect of the chosen uncertainty measure. Numerous different measures are used across the literature, for example predictive and expected entropy, mutual information and variance, and it is not clear which measure best reflects uncertainty.

This paper makes the following contributions:
\begin{itemize}
    \item We introduce a new Bayesian parameter estimation method called RECAST (\textbf{re}start \textbf{c}osine \textbf{a}nnealing \textbf{st}ochastic gradient Langevin dynamics). It is a new method of SG-MCMC that enables exploration of more complex posterior distributions for neural network parameters.
    \item  We conduct a principled comparison of different uncertainty measures that results in a calibrated method to estimate uncertainty.
    \item When benchmarked using mutilated data, we show that predictive entropy and epistemic uncertainty when combined with RECAST offers the best calibrated measure of uncertainty when compared to recent methods.   
\end{itemize}

\section{Existing Methods for Uncertainty Estimation in Bayesian Deep Learning}

In conventional neural networks, the parameters are estimated by a single point value obatained using backpropagation with stochastic gradient descent (SGD) ~\cite{corr/abs-1811-07969}. In contrast, Bayesian Neural Networks (BNNs) assume a prior over model parameters $\theta$ and then data is used to compute a distribution over each of these parameters. During training, the data is used to update the posterior distribution ($P(\theta|x,y)$) over the BNN's parameters, using Bayes rule in Equation 1. \\
\begin{equation} \label{bayes}
P(\theta|x,y)=\frac{P(x,y|\theta)P(\theta)}{\int P(y|x,\theta)P(\theta)d\theta }
\end{equation}\\
\noindent Where: \\
\begin{tabular}{ll}
	$\theta$  & = Neural networks parameters \\
	$x $ & = Networks input\\
	$ y $ & =  Networks output \\
\end{tabular}

However, for BNNs with thousands of parameters, computing the posterior is intractable due to the complexity in computing the marginal likelihood~\cite{ye2019functional}. Markov Chain Monte Carlo (MCMC) or Variational Inference (VI) methods are a solution to the complexity. The trade-off is that MCMC has a higher variance and lower bias in the estimate, while VI has a higher bias but lower variance~\cite{mattei2019parsimonious}.

In the following subsections we review work related to uncertainty estimation in BNNs and how to train these models. 

\subsection{Markov Chain Monte Carlo (MCMC)}

The gold standard for Bayesian inference has typically been MCMC, which iteratively draws samples from an unknown posterior distribution. For neural networks, both Metropolis-Hastings and Gibbs sampling have been shown to be computationally intractable even for smaller neural networks. 

Hamiltonian Monte Carlo (HMC) \cite{neal2011mcmc} addresses part of the computational challenge associated with MCMC by making use of gradient information as opposed to random walks from the aforementioned methods. However, it still fails to scale to deeper networks by computing gradients from the entire dataset, resulting in a computation complexity of $\mathcal{O}(n)$, where n is the dataset size. 

By using mini-batches for the gradient computation~\cite{welling2011bayesian}, Stochastic Gradient (SG)-MCMC mitigates the need to compute gradients using the full dataset. This approach enables easier computation (with the same computational complexity as SGD) namely $\mathcal{O}(m)$, where m is the mini-batch size. While, SG-MCMC has been widely used for BNNs~\cite{li,ye2019functional,welling2011bayesian,park,maddoxfast}, the drawback is the inability to capture complex distributions in the parameter space, without increasing the computational overhead.

\subsection{Variational Inference}

Variational inference performs Bayesian inference by using a computationally tractable "variational" distribution to approximate the posterior~\cite{ye2019functional}. The goal is to minimize the Kullback-Leibler (KL) divergence between the tractable variational distribution ($q(\theta)$), which is typically a Gaussian, and the true posterior ($p(\theta|x,y$))~\cite{gal2015dropout}, i.e
\begin{equation}\label{kl}
KL(q(\theta)|p(\theta|\textbf{X,Y})).
\end{equation}
\noindent Where: \\
\begin{tabular}{ll}
$q(\theta)$ & = variational distribution \\
$p(\theta|x,y$) & =  True posterior \\\\	
\end{tabular}

This minimization is equivalent to maximization of the log evidence lower bound (ELBO), i.e., \\
\begin{equation}\label{loge}
L = \int q(\theta)log p(\textbf{Y}|\textbf{X},\theta)d\theta - KL(q(\theta)|p(\theta)).
\end{equation}
\noindent Where: \\
\begin{tabular}{ll}
	$\int q(\theta)log p(\textbf{Y}|\textbf{X},\theta)d\theta $  & = Likelihood wrt $q(\theta)$\\
	$KL(q(\theta)|p(\theta))$ & = KL divergence between the variational distribution and prior\\	\\
\end{tabular}

Multiple variational inference methods have been proposed as the alternative to MCMC. Graves~\cite{graves2011practical} proposed that when estimating the weights of neural networks, a Gaussian variational posterior can be used to approximate the distribution of the weights in a network. Despite having lower computational overhead, when applied for uncertainty estimation in BNNs, the capacity of the uncertainty representation is limited by the variational distribution $q(\theta)$. 

Gal and Ghahramani~\cite{gal2015dropout} showed that variational inference can be approximated without modifying the network. This is achieved through a method of approximate variational inference called Monte Carlo Dropout (MCD), whereby dropout is performed at test time, using multiple dropout masks.

\subsection{Frequentist Approximations}
Bootstrapping and ensembling are frequentist methods that can be used to estimate neural network uncertainty without the Bayesian computational overhead as well as being easily parallelizable \cite{lakshminarayanan2017simple}. Bootstrapping makes multiple random draws from the training data with replacement~\cite{DBLP:books/lib/HastieTF09}. On each draw, the model parameters are estimated. However, as shown by Lakshminarayanan et al \cite{lakshminarayanan2017simple} bootstrapping leads to degraded performance. Hence, Deep Ensembles \cite{lakshminarayanan2017simple} was proposed as a solution by training multiple randomly initialized neural networks. At test time the output variance from the ensemble of models is used as an estimate of uncertainty~\cite{lakshminarayanan2017simple}.

\section{Our Approach: RECAST}

RECAST is based on Stochastic Gradient Langevin Dynamics (SGLD)~\cite{welling2011bayesian}, with the goal of improving its ability to perform uncertainty estimation, whilst ensuring scalability by retaining the computational complexity of $\mathcal{O}(m)$. Before discussing RECAST, we review SGLD.

SGLD is a gradient based MCMC algorithm, which enables faster sampling from the posterior than HMC by computing gradients using mini-batches. SGLD has two steps:
\begin{enumerate}
	\item \textit{Stochastic optimization step}: Update the maximum a-posteriori (MAP) estimate on each mini batch of size m. 
	\item \textit{Langevin dynamics phase}: over N iterations evaluate the gradient steps with a decreasing step size ($\epsilon_{t}$) and add Gaussian noise ($\eta_{t}$). 

\end{enumerate}

	Equation \ref{langevin} characterizes the process for N forward passes, where $x_{ti}$ is the ith minibatch of data. It illustrates that as the step size ($\epsilon_{t}$)  decays to zero, the noise term ($\eta_{t}$) begins to dominate. The method then approximates Langevin Monte Carlo to sample from the posterior over parameters.

\begin{equation}\label{langevin}
\Delta \theta_{t} = \frac{\epsilon_{t}}{2} (\nabla log p(\theta_{t}) + \frac{N}{n}\sum_{1}^{n} \nabla log p(x_{ti}|\theta_{t} )) + \eta_{t}
\end{equation}

\noindent Where \\
\begin{tabular}{llll}
  $\theta$ & = Neural network parameter vector & $p(\theta) $ & = Prior distribution \\
   	N  & = Dataset size &  n &  = subset of N (s.t. n < N) \\
   $p(x|\theta)$ & = Likelihood of data given parameters & $\epsilon_{t}$ & = Step size at t \\
   $ \eta_{t}$ & = $\sim \mathcal{N}(0,\,\epsilon_{t})$ &
	$x_{ti}$ & = ith mini-batch of data\\
	\\	
\end{tabular}

\begin{algorithm}[h]
   \caption{The RECAST algorithm \label{alg:recast}}
\begin{algorithmic}
   \STATE {\bfseries Input:} Step size ($\epsilon_{t}$), Number of samples (N), Restart Iterations (R)
   \STATE Initialize $\epsilon_{t}$ = 1.
   \FOR{$i=1$ {\bfseries to} N}
   \IF{$i~mod~R=0$ (i.e. every R iterations)}
   \STATE  \textit{Reset and Exploration stage:\\} Apply the warm restart by resetting the step size ($\epsilon_{t}$) to 1. The large $\epsilon_{t}$ allows for an exploration of the posterior.
   \ENDIF
   \STATE \textit{Step size decay and Exploitation stage:\\} Decay step size ($\epsilon_{t}$) according to a cosine annealed schedule. During this phase (for small $\epsilon_{t}$) samples are drawn from the localized area of the posterior.
   \ENDFOR
\end{algorithmic}
\end{algorithm}
\newpage
The caveat is that for SGLD posterior sampling, the step size ($\epsilon_{t}$) must satisfy the following:
    \begin{center}
         (1) $\sum_{t=1}^{\infty}\epsilon_{t}=\infty$ and (2)
    $\sum_{t=1}^{\infty}\epsilon_{t}^{2}<\infty$
    \end{center}

Ultimately, instead of
training a single network, SG-MCMC trains an ensemble of
networks, where each network has its weights drawn from
a shared posterior distribution. The average of the samples is simply the expectation of the posterior over the model parameters  ($\theta$).

SGLD's convergence properties are related to the step size ($\epsilon_{t}$). However, as the step size decays (during the Langevin dynamics phase), the sampling efficiency also decreases. This results in sampling within a localized area which reduces the capacity to explore the posterior.  

RECAST addresses the issue of sampling efficiency in SGLD, whilst not increasing the computational cost ($\mathcal{O}(m)$). The method draws from learning rates schedules with cosine annealling and warm restarts used in stochastic gradient descent (SGD) \cite{Loshchilov2017SGDRSG}. RECAST is based on the hypothesis that the singular SGLD step size decay is unable to sufficiently explore more complex posterior distributions. 

Importantly when Bayesian methods fail to capture the true posterior it results in models being miscalibrated \cite{kuleshov2018accurate}. This is discussed further in our experimental evaluation in Section 5. However, RECAST presented in Algorithm 1, overcomes this issue by introducing warm restarts to the SGLD learning rate ($\epsilon_{t}$) and can explore a more diverse posterior over parameters. 
We present an overview of RECAST in Algorithm \ref{alg:recast}, specifically noting that that warm restarts introduce two phases.

We term the first phase the exploitation phase which mirrors SGLD and samples a localized area of the posterior. The warm restarts then introduce the second phase termed the Reset and Exploration phase. During the periods of sampling inefficiency the warm restart is applied such that the learning rate is reset back to a larger value. This allows for a larger step to be taken, leading to exploration and sampling of other areas of the posterior. The process can be thought of as analogous to a conditioned exploration vs exploitation trade-off.

\section{Uncertainty Measures and Calibration}
\subsection{Uncertainty Measures}
Many works overlook how models' perform for different uncertainty measures. To address this, we compare four widely used uncertainty measures: predictive entropy, mutual information, aleatoric uncertainty and epistemic uncertainty. We review the different measures of uncertainty next. 

\begin{itemize}
    \item \textit{Predictive Entropy:} which is widely used in information theory
has been argued to be a good measure to evaluate uncertainty~\cite{Leibig2017,lakshminarayanan2017simple}, where a higher predictive entropy corresponds to a greater degree of uncertainty \cite{mackay2003information}. This measure is given by:
\begin{equation}\label{exent}
    H=-\sum_{y \in Y}^{} P(y|x)logP(y|x)
\end{equation}
\noindent Where: \\
\begin{tabular}{ll}
	$P(y|x)$  & = softmax output of the network
\end{tabular}

\item \textit{Mutual Information:}
 is the information gain related to the model parameters for the dataset if we see a label $y$ for an input $x$. It is  predictive entropy minus the expected entropy. i.e.,
\begin{equation}\label{mi}
    I = H[P(y | x,D)]-\mathbb{E}_{p(w|D)}H[P(y|x,w)].
\end{equation}
\noindent Where: \\
\begin{tabular}{ll}
	$H[P(y | x,D)]$  & = Predictive entropy \\
	$\mathbb{E}_{p(w|D)}H[P(y|x,w)]$ & = Expected entropy\
\end{tabular}
 
\item \textit{Aleatoric Uncertainty:}
 captures the inherent noise/stochasticity in the data \cite{Kendall2017,sri,Gal2017}. Hence, increasing the dataset size will not impact uncertainty. Thus, for mutilated data the uncertainty should be high. Aleatoric uncertainty is calculated as per \cite{kwon2018uncertainty}
\begin{equation}\label{alea}
   \frac{1}{T}\sum_{t=1}^{T}diag(\hat{p_{t}})-\hat{p_{t}}\hat{p_{t}}^{T}
\end{equation}
\noindent Where: \\
\begin{tabular}{ll}
	$\hat{p_{t}} $  & =  softmax $(f_{w_{t}}(x^{*})$) 
\end{tabular}

\item \textit{Epistemic Uncertainty:}
is the inherent model uncertainty \cite{Kendall2017,sri,Gal2017}. When inputs are similar to the training data there will be lower uncertainty, whilst data that is different from the original training data (such as mutilated data or out of distribution) should have a higher epistemic uncertainty \cite{Kendall2017,sri,Gal2017}. Epistemic uncertainty is given by 
\begin{equation}\label{epi}
   \frac{1}{T}\sum_{t=1}^{T}(\hat{p_{t}}-\bar{p_{t}})(\hat{p_{t}}-\bar{p_{t}})^{T}
\end{equation}
\noindent Where: \\
\begin{tabular}{ll}
	$ \bar{p_{t}}$  = $\frac{1}{T}\sum_{t=1}^{T}\hat{p_{t}}$ 
\end{tabular}
\end{itemize}

\subsection{Uncertainty Calibration}
Calibration is essential so that predictive probabilities from the softmax are useful as confidence measures. We define calibration as the probability that a predicted class label ($\hat{Y} $) reflects the ground truth likelihood \cite{guo2017calibration}. 
For example for a well calibrated model, if a class label is assigned with predictive probability of 0.8, we expect this to be true 80\% of the time \cite{guo2017calibration}.  Expected Calibration Error (ECE) is a widely used baseline for calibration and is given by Equation \ref{ECE}.
\begin{equation}\label{ECE}
ECE = \mathbb{E}[|\mathbb{P}(\hat{Y}=Y | \hat{P}=p) - p|] \newline
 = \sum_{m=1}^{M} \frac{|B_{m}|}{n} |acc(B_{m})-conf(B_{m})| 
\end{equation}
\noindent Where: \\
\begin{tabular}{llrl}
	$n$  & = number of samples in the bin & $M$ & = number of bins\\
	$acc$ & = average accuracy for bin $B_{m}$ &
	$conf$  & = confidence for bin $B_{m}$
\end{tabular}

Hence perfect model calibration would have an ECE equal to 0. However, it is impossible to achieve perfect calibration even on unmutilated data \cite{guo2017calibration}. This is more so true when the data becomes mutilated (i.e. simulated uncertainty). Thus, we propose an addition to our definition. The model remains calibrated if the increase in ECE and the uncertainty measures correlate with the increase in mutilation (added uncertainty). A perfect calibration under mutilation would have a correlation equaling 1.

\section{Experimental evaluation}
We analyze the effectiveness of the following uncertainty measures: aleatoric uncertainty, epistemic uncertainty, mutual information and expected entropy using different parameter estimation methods including our approach RECAST, as well as SGLD \cite{welling2011bayesian}, MCD \cite{gal2015dropout}, Variational Inference (VI) \cite{blundell2015weight}  and Deep Ensembling (ENSEMBLE) \cite{lakshminarayanan2017simple}.
Additionally, we evaluate calibration under uncertainty based on the expected calibration error (ECE). We compare the methods on benchmark image classification datasets of MNIST and CIFAR-10. Training uses un-mutilated images for each BNN parameter estimation method and testing uses mutilated images to simulate uncertainty.

The images undergo two different mutilations to simulate uncertainty: noise and rotation mutilations. After inference using the mutilated images, the uncertainty measures are then computed for the different BNN parameter estimation methods. Finally, we assess the expected calibration error (ECE), with 15 bins.  Furthermore, based on our definition of calibrated uncertainty, we evaluate both ECE and the uncertainty measures for increasing mutilation. 

As discussed perfect calibration would have zero ECE, however it is impossible to achieve even with un-mutilated data \cite{guo2017calibration}. Therefore,  we assess the correlation between the ECE and uncertainty measure, which for perfect calibration under uncertainty should equal 1. This implies that the model remains calibrated under the uncertainty and that the increase in ECE is only a function of the mutilation with the model retaining it's latent calibration. We additionally evaluate the correlation between the uncertainty measure and model accuracy. Ideal performance would be a correlation of 1, where the uncertainty measure directly reflects the accuracy of the prediction. The cosine similarity between the uncertainty measure and mutilation value was also computed. However, we omit the cosine similarity result as it had similar performance to the correlation values listed in Table 1. By similar performance we refer to both the magnitudes of similarity and relative rankings of the different methods.

\begin{figure}[t]
    \centering
    \includegraphics[width=\textwidth]{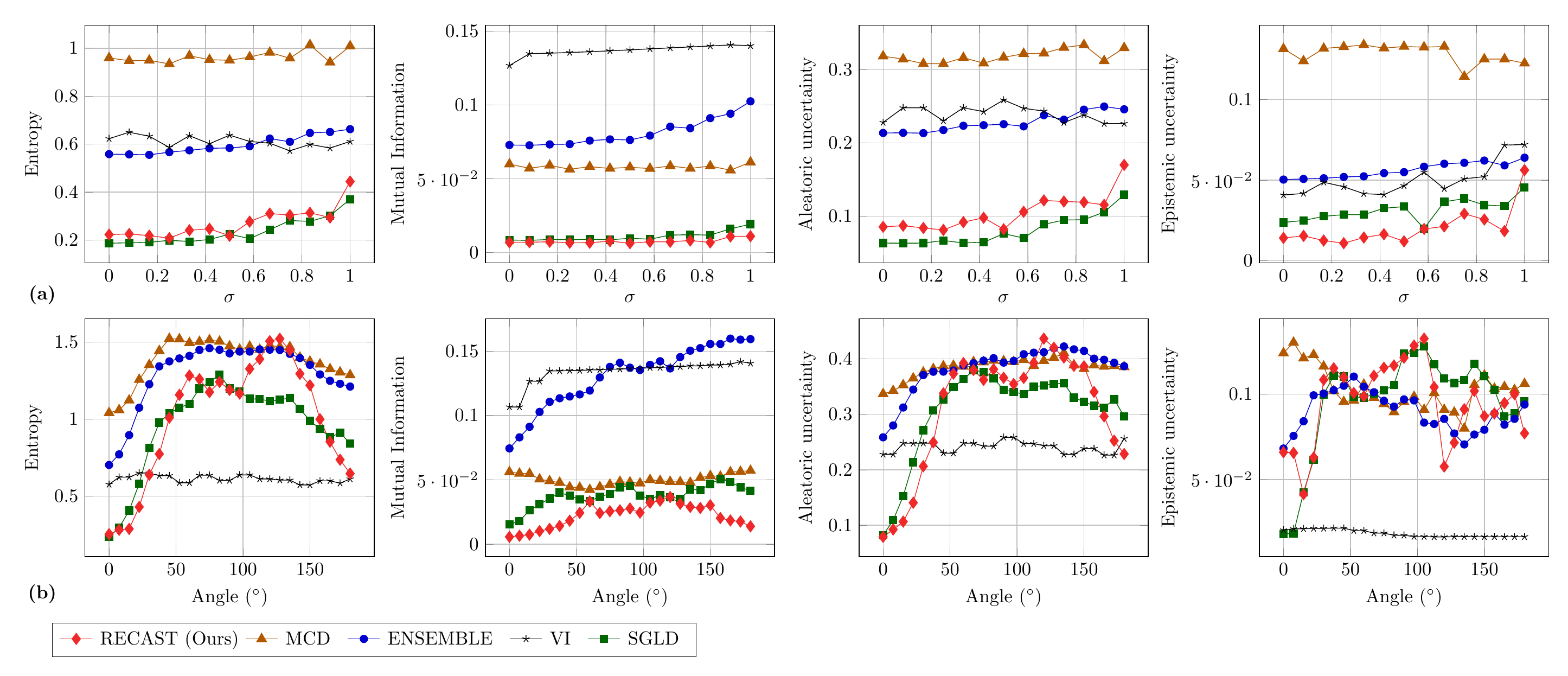}
    \caption{Uncertainty measures computed for the different BNN parameter estimation methods with increasing levels of image mutilation by (a) Gaussian noise ($\sigma$) and (b) Rotation (\textdegree)}
    \label{fig:plots}
\end{figure}

\subsection{Comparison Methods}
RECAST is shown to have a high performance which is discussed in detail in Section 5.2. Empirically it was found that this best performance was achieved for warm restarts after 2000 iterations. We propose that restarts after fewer iterations result in not reaching the Langevin dynamics phase and a greater number of iterations results in lower diversity captured in the posterior. In addition to RECAST, we compare the following approaches to parameter estimation in BNNs:
\begin{enumerate}
    \item MCD -- Monte Carlo (MC) dropout~\cite{gal2015dropout} involves doing N Monte Carlo samples (i.e. Infer $y|x$ N times). At each inference iteration, sample a different set of units to drop out. This generates random predictions, which are interpreted as samples from a probabilistic distribution \cite{gal2015dropout}.
    \item ENSEMBLE \cite{lakshminarayanan2017simple} -- Twelve randomly initialized point estimate networks are trained  \cite{lakshminarayanan2017simple}, thereby ensuring multiple different estimates of the model parameters. After performing inference using each model, the uncertainty measures are combined as an expected value. 
    \item VI -- Variational Inference is implemented using Bayes by Backprop \cite{blundell2015weight}. The variational distribution that minimizes the KL divergence is estimated through sampling at each
iteration of backpropagation. Specifically, each of the likelihoods are parameterized by the variational distribution $q(\textbf{w}| \theta)$. Thus, each likelihood is a estimated by sample from $q(\textbf{w}| \theta)$. 
    \item SGLD -- We compare RECAST to standard SGLD. Hence, we use the same number of sample iterations as RECAST. However, the step size decays as originally proposed \cite{welling2011bayesian}, with $\epsilon_{t}=a(b+t)^{-0.55}$, where a and b are 1 and 1 so that $\epsilon_{t}$ decays from 1 to 0.0057.
\end{enumerate}

\subsection{Uncertainty Evaluation}

We demonstrate the effectiveness of SG-MCMC based methods applied to neural networks. Specifically, highlighting the effectiveness of RECAST over the other methods benchmarked. Evaluation is conducted using a LeNet-5 CNN with a Rectified Linear Unit (ReLu) activation function. A maximum of 12000 epochs was used for all methods to ensure adequate sampling of the distributions. All experiments are carried out as follows and repeated five times for stability:

\begin{enumerate}
    \item Train different BNNs using the different parameter estimation methods. The training is conducted on the original, un-mutilated images.
    \item  At test time uncertainty is simulated by applying either the noise or the rotation mutilation. (i) The noise image mutilation involves adding a Gaussian noise kernel (variance between 0-1, in 0.083 increments). 
    (ii) The rotation mutilation rotates the image between\\ 0-180\textdegree~in 7.5\textdegree~increments.
    \item After each mutilation iteration is applied, inference is performed using the trained network and the four uncertainty measures are computed for the sample.
    \item Repeat 1-3 for all test examples. Average the results per discrete mutilation.
\end{enumerate}

The results comparing different uncertainty measures for each implemented parameter estimation method is shown in Figures 1 (a) and (b) respectively. Two important factors are captured in both the results. Firstly, the performance for the original images (i.e. no mutilation: sigma and angle of zero). Secondly, the performance relative to the simulated uncertainty by mutilation. 

We show that RECAST and SGLD best capture uncertainty as they scale from no mutilation to higher uncertainty for greater mutilation. VI and ENSEMBLE convey similar uncertainty for all uncertainty measures beside epistemic uncertainty. Finally MCD shows high values of uncertainty throughout, even with no mutilation applied. This is highly indicative of miscalibration.

The relative performance of the methods is then quantified in Table 1. As discussed, correlation is used as a metric for the calibration mapping, where a higher magnitude represents better calibration. We evaluate the correlation of ECE with uncertainty, as well as, presenting the overall model calibration based on ECE.

\begin{table}[t]
\centering
\caption{Comparison of BNN parameter estimation methods. (1) The correlation of uncertainty measures and accuracy, (2) correlation of ECE and uncertainty and (3) model calibration based on ECE is presented. }
\label{sample-table}
\vskip 0.15in
\begin{sc}
\scalebox{0.85}{\begin{tabular}{lccccc}
\toprule
\textit{1.Uncertainty vs Acc Corr} & \textbf{RECAST (Ours)} & MCD & ENSEMBLE  & VI & SGLD    \\

\midrule

\textbf{Entropy}   & \textbf{0.84} $\pm$ \textbf{0.08}  & 0.28 $\pm$ 0.12 & 0.70 $\pm$ 0.13  & 0.65 $\pm$ 0.14 & 0.84 $\pm$ 0.11 \\
\textbf{Epistemic Uncertainty}  & \textbf{0.83} $\pm$ \textbf{0.09} & 0.13 $\pm$ 0.09    & 0.53 $\pm$ 0.17  & 0.75 $\pm$ 0.14  & 0.57 $\pm$ 0.11 \\
Mutual Information   & 0.78 $\pm$ 0.07 & 0.32 $\pm$ 0.19   & 0.65 $\pm$ 0.5  & 0.65 $\pm$ 0.02  & 0.91 $\pm$ 0.05  \\
Aleatoric Uncertainty   & 0.83 $\pm$ 0.09 & 0.54 $\pm$ 0.05   & 0.83 $\pm$ 0.10  & 0.78 $\pm$ 0.06  & 0.18 $\pm$ 0.12  \\
\bottomrule
\textit{2.ECE vs Uncertainty Corr}  &\textbf{0.91} $\pm$ \textbf{0.04} &  0.51 $\pm$ 0.11 & 0.66 $\pm$ 0.12  & 0.61 $\pm$ 0.05  & 0.61 $\pm$ 0.09 \\
\bottomrule
\textit{3.Model ECE}  & \textbf{2.37}  & 13.99 & 3.93 & 8.60 & 3.97 \\
\bottomrule
\end{tabular}}
\end{sc}
\vskip -0.1in
\end{table}

The strength of RECAST is noted based on the correlation between the uncertainty measure and the model accuracy. Moreover, RECAST allows the model to retain calibration under uncertainty based on the significantly lower ECE when compared to all other methods. In addition, the higher correlation of ECE to uncertainty indicates RECAST scales best to the uncertainty.

When evaluating all the uncertainty measures, it is evident that predictive entropy and epistemic uncertainty provide the best mapping of uncertainty to mutilations, irrespective of parameter estimation method. However, when comparing parameter estimation methods, MCD has the poorest correlation magnitudes, whilst VI and ENSEMBLE also match up equally.

While RECAST and SGLD have similar performance, a key distinction lies in epistemic uncertainty. RECAST has a superior correlation between the epistemic uncertainty and the model accuracy, when compared to SGLD. This not only represents the superior calibration of RECAST, but it is also useful as  measuring epistemic uncertainty would allow the model accuracy to be determined by proxy.

In particular, RECAST highlights that high epistemic uncertainty is linked to poor performance in classification. This proxy useful for cases where at test time the ground truth labels are not readily available and yet we wish to infer classification performance of the deployed model.

In terms of model calibration, RECAST provides the best calibration with the lowest ECE of 2.37\%. We highlight as well, that RECAST  provides a better calibration based on ECE when compared to a baseline from \cite{guo2017calibration} who also used a LeNet-5 on the same dataset with a 3.02\% ECE.

RECAST further outperforms all models based on the correlation of ECE vs uncertainty. This shows RECAST better retains its level of calibration behaviour even under mutilations, with ECE increasing as a linear dependant function of mutilation.  

Overall, we show that RECAST is better calibrated than other methods for uncertainty estimation when using entropy or epistemic uncertainty measures, whilst still being scalable by retaining the complexity of $\mathcal{O}(m)$.

\subsection{Weight Distribution Analysis}

\begin{wrapfigure}{r}{0.4\textwidth}
\vspace{-2.0cm}
  \begin{center}
   \includegraphics[width=\linewidth]{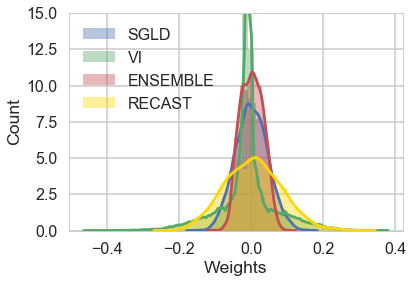}
  \end{center}
  \caption{Distributions of weights over parameters for the BNNs.\label{fig:wdist}}
\end{wrapfigure}

To examine if RECAST offers a greater exploration of the posterior and captures a more diverse weight space, we sampled the weights in all of the BNNs, which is shown in Figure \ref{fig:wdist}. RECAST (gold) exhibits a greater diversity of weights than other BNNs. 

This diversity helps explain RECAST's better ability to represent uncertainty compared to the other BNNs. Moreover, offering an explanation as to why RECAST retains calibration even under uncertainty. A narrower distribution over parameters implies insufficient model capacity, which in turn causes poor performance in representing uncertainty.  Moreover, our results which demonstrate poor calibration for the BNNs with narrower distributions, validate that failing to capture the full posterior can result in miscalibration.

\section{Discussion}
This paper investigated key aspects of calibrated and scalable uncertainty representations for neural networks. We have highlighted that SG-MCMC algorithms provide the best uncertainty representations for neural networks, when tested with mutilated inputs. Our proposed method, RECAST achieves the best performance against all other benchmarked methods of parameter estimation, whilst retaining the computational complexity of $\mathcal{O}(m)$ for scalability.

In particular, we showed that the uncertainty represented by RECAST has the best calibration based on the ECE and also as indicated by the high correlations when compared to the other methods. This is highly relevant for sensitive applications presented at the beginning of the paper. Specifically, we can use the uncertainty measured for models employing RECAST as a proxy for the model's accuracy of prediction, whereby a high uncertainty is likely to result in a poor prediction by the neural network.

Moreover, we have shown that RECAST, which combines cosine annealling with warm restarts with SGLD, captures a more diverse representation of the parameters in the network. Hence, the increased diversity increases the model's ability to represent uncertainty.

Besides, comparing performance of different methods of parameter estimation, this work is also unique by analyzing how the measure of uncertainty affects the actual representation of the uncertainty. We have shown that predictive entropy and epistemic uncertainty provide the most robust representations both for different models, but also for different mutilations. 

Ultimately, we have shown that the representation of uncertainty in neural networks provides greater robustness to predictions. We compared parameter estimation methods for neural networks and demonstrated that our method RECAST offers richer and more diverse representations over parameters. When RECAST is paired with entropy and epistemic uncertainty, it yielded a well-calibrated measure of uncertainty that scales to neural networks.


{\small
\bibliographystyle{ieee}
\bibliography{refs}

\begin{thebibliography}{10}\itemsep=-1pt

\bibitem{blundell2015weight}
C.~Blundell, J.~Cornebise, K.~Kavukcuoglu, and D.~Wierstra.
\newblock Weight uncertainty in neural networks.
\newblock {\em arXiv preprint arXiv:1505.05424}, 2015.

\bibitem{gal2015dropout}
Y.~Gal and Z.~Ghahramani.
\newblock Dropout as a bayesian approximation: Insights and applications.
\newblock In {\em Deep Learning Workshop, ICML}, volume~1, page~2, 2015.

\bibitem{Gal2017}
Y.~Gal, J.~Hron, and A.~Kendall.
\newblock Concrete dropout.
\newblock In I.~Guyon, U.~V. Luxburg, S.~Bengio, H.~Wallach, R.~Fergus,
  S.~Vishwanathan, and R.~Garnett, editors, {\em Advances in Neural Information
  Processing Systems 30}, pages 3581--3590. Curran Associates, Inc., 2017.

\bibitem{graves2011practical}
A.~Graves.
\newblock Practical variational inference for neural networks.
\newblock In {\em Advances in neural information processing systems}, pages
  2348--2356, 2011.

\bibitem{speech}
A.~{Graves}, A.~{Mohamed}, and G.~{Hinton}.
\newblock Speech recognition with deep recurrent neural networks.
\newblock In {\em 2013 IEEE International Conference on Acoustics, Speech and
  Signal Processing}, pages 6645--6649, May 2013.

\bibitem{guo2017calibration}
C.~Guo, G.~Pleiss, Y.~Sun, and K.~Q. Weinberger.
\newblock On calibration of modern neural networks.
\newblock In {\em Proceedings of the 34th International Conference on Machine
  Learning}, pages 1321--1330. JMLR. org, 2017.

\bibitem{DBLP:books/lib/HastieTF09}
T.~Hastie, R.~Tibshirani, and J.~H. Friedman.
\newblock {\em The elements of statistical learning: data mining, inference,
  and prediction, 2nd Edition}.
\newblock Springer series in statistics. Springer, 2009.

\bibitem{Kendall2017}
A.~Kendall and Y.~Gal.
\newblock What uncertainties do we need in bayesian deep learning for computer
  vision?
\newblock In I.~Guyon, U.~V. Luxburg, S.~Bengio, H.~Wallach, R.~Fergus,
  S.~Vishwanathan, and R.~Garnett, editors, {\em Advances in Neural Information
  Processing Systems 30}, pages 5574--5584. Curran Associates, Inc., 2017.

\bibitem{corr/abs-1811-07969}
A.~Kortylewski, M.~Wieser, A.~Morel{-}Forster, A.~Wieczorek, S.~Parbhoo,
  V.~Roth, and T.~Vetter.
\newblock Informed mcmcwith bayesian neural networks for facial image analysis.
\newblock {\em Bayesian Deep Learning Workshop at NeurIPS 2018},
  abs/1811.07969, 2018.

\bibitem{Krizhevsky2012}
A.~Krizhevsky, I.~Sutskever, and G.~E. Hinton.
\newblock Imagenet classification with deep convolutional neural networks.
\newblock In F.~Pereira, C.~J.~C. Burges, L.~Bottou, and K.~Q. Weinberger,
  editors, {\em Advances in Neural Information Processing Systems 25}, pages
  1097--1105. Curran Associates, Inc., 2012.

\bibitem{kuleshov2018accurate}
V.~Kuleshov, N.~Fenner, and S.~Ermon.
\newblock Accurate uncertainties for deep learning using calibrated regression.
\newblock {\em arXiv preprint arXiv:1807.00263}, 2018.

\bibitem{kwon2018uncertainty}
Y.~Kwon, J.-H. Won, B.~J. Kim, and M.~C. Paik.
\newblock Uncertainty quantification using bayesian neural networks in
  classification: Application to ischemic stroke lesion segmentation.
\newblock 2018.

\bibitem{lakshminarayanan2017simple}
B.~Lakshminarayanan, A.~Pritzel, and C.~Blundell.
\newblock Simple and scalable predictive uncertainty estimation using deep
  ensembles.
\newblock In {\em Advances in Neural Information Processing Systems}, pages
  6402--6413, 2017.

\bibitem{Leibig2017}
C.~Leibig, V.~Allken, M.~S. Ayhan, P.~Berens, and S.~Wahl.
\newblock Leveraging uncertainty information from deep neural networks for
  disease detection.
\newblock {\em Scientific Reports}, 7(1):17816, Dec. 2017.

\bibitem{li}
C.~{Li}, A.~{Stevens}, C.~{Chen}, Y.~{Pu}, Z.~{Gan}, and L.~{Carin}.
\newblock Learning weight uncertainty with stochastic gradient mcmc for shape
  classification.
\newblock In {\em 2016 IEEE Conference on Computer Vision and Pattern
  Recognition (CVPR)}, pages 5666--5675, June 2016.

\bibitem{medical}
G.~Litjens, T.~Kooi, B.~E. Bejnordi, A.~A.~A. Setio, F.~Ciompi, M.~Ghafoorian,
  J.~A. Van Der~Laak, B.~Van~Ginneken, and C.~I. S{\'a}nchez.
\newblock A survey on deep learning in medical image analysis.
\newblock {\em Medical image analysis}, 42:60--88, 2017.

\bibitem{Loshchilov2017SGDRSG}
I.~Loshchilov and F.~Hutter.
\newblock Sgdr: Stochastic gradient descent with warm restarts.
\newblock {\em ICLR}, 2017.

\bibitem{mackay2003information}
D.~J. MacKay.
\newblock {\em Information theory, inference and learning algorithms}.
\newblock Cambridge university press, 2003.

\bibitem{maddoxfast}
W.~Maddox, T.~Garipov, P.~Izmailov, D.~Vetrov, and A.~G. Wilson.
\newblock A simple baseline for bayesian uncertainty in deep learning.
\newblock {\em arXiv preprint arXiv:1902.02476}, 2019.

\bibitem{mattei2019parsimonious}
P.-A. Mattei.
\newblock A parsimonious tour of bayesian model uncertainty.
\newblock {\em arXiv preprint arXiv:1902.05539}, 2019.

\bibitem{neal2011mcmc}
R.~M. Neal et~al.
\newblock Mcmc using hamiltonian dynamics.
\newblock {\em Handbook of markov chain monte carlo}, 2(11):2, 2011.

\bibitem{nguyen2015deep}
A.~Nguyen, J.~Yosinski, and J.~Clune.
\newblock Deep neural networks are easily fooled: High confidence predictions
  for unrecognizable images.
\newblock In {\em Proceedings of the IEEE conference on computer vision and
  pattern recognition}, pages 427--436, 2015.

\bibitem{park}
C.~Park, J.~Kim, S.~H. Ha, and J.~Lee.
\newblock Sampling-based bayesian inference with gradient uncertainty.
\newblock {\em Workshop on Bayesian Deep Learning, NeurIPS 2018},
  abs/1812.03285, 2018.

\bibitem{Pearce2018UncertaintyIN}
T.~Pearce, M.~Zaki, A.~Brintrup, and A.~Neely.
\newblock Uncertainty in neural networks: Bayesian ensembling.
\newblock {\em CoRR}, abs/1810.05546, 2018.

\bibitem{car}
Q.~Rao and J.~Frtunikj.
\newblock Deep learning for self-driving cars: Chances and challenges.
\newblock In {\em Proceedings of the 1st International Workshop on Software
  Engineering for AI in Autonomous Systems}, SEFAIS '18, pages 35--38, New
  York, NY, USA, 2018. ACM.

\bibitem{sri}
K.~Shridhar, F.~Laumann, and M.~Liwicki.
\newblock A comprehensive guide to bayesian convolutional neural network with
  variational inference.
\newblock {\em CoRR}, abs/1901.02731, 2019.

\bibitem{welling2011bayesian}
M.~Welling and Y.~W. Teh.
\newblock Bayesian learning via stochastic gradient langevin dynamics.
\newblock In {\em Proceedings of the 28th international conference on machine
  learning (ICML-11)}, pages 681--688, 2011.

\bibitem{ye2019functional}
N.~Ye and Z.~Zhu.
\newblock Functional bayesian neural networks for model uncertainty
  quantification, 2019.

\end{thebibliography}
}


\end{document}